\documentclass[runningheads]{llncs}
\usepackage[T1]{fontenc}
\usepackage{graphicx}
\usepackage{hyperref}
\usepackage{color}

\usepackage{float}
\usepackage[frozencache=true]{minted}
\begin{document}
\title{\texttt{gym-saturation}: Gymnasium environments for saturation provers (System description) \thanks{This work has been supported by the French government, through the
3IA Côte d’Azur Investment in the Future project managed by the National Research Agency (ANR) with the reference numbers ANR-19-P3IA-0002.}}
\titlerunning{\texttt{gym-saturation}: Gymnasium environments for saturation provers}
%
\author{Boris Shminke\orcidID{0000-0002-1291-9896}}
\authorrunning{B. Shminke}
%
\institute{Université Côte d’Azur, CNRS, LJAD, France \\
\email{boris.shminke@univ-cotedazur.fr}}
\maketitle              
\begin{abstract}
This work describes a new version of a previously published Python package --- \texttt{gym-saturation}: a collection of OpenAI Gym environments for guiding saturation-style provers based on the given clause algorithm with reinforcement learning. We contribute usage examples with two different provers: Vampire and iProver. We also have decoupled the proof state representation from reinforcement learning per se and provided examples of using a known \texttt{ast2vec} Python code embedding model as a first-order logic representation. In addition, we demonstrate how environment wrappers can transform a prover into a problem similar to a multi-armed bandit. We applied two reinforcement learning algorithms (Thompson sampling and Proximal policy optimisation) implemented in Ray RLlib to show the ease of experimentation with the new release of our package.

\end{abstract}
\section{Introduction}
This work describes a new version (\texttt{0.10.0}, released 2023.04.25) of a previously published~\cite{Shminke2022} Python package --- \texttt{gym-saturation}~\footnote{\url{https://pypi.org/project/gym-saturation/}}: a collection of OpenAI~Gym~\cite{DBLP:journals/corr/BrockmanCPSSTZ16} environments for guiding saturation-style provers (using the given clause algorithm) with reinforcement learning (RL) algorithms. The new version partly implements the ideas of our project proposal~\cite{https://doi.org/10.48550/arxiv.2209.02562}. The main changes from the previous release (\texttt{0.2.9}, on 2022.02.26) are:
\begin{itemize}
\item guiding two popular provers instead of a single experimental one (Section~\ref{section:implementation})
\item pluggable first-order logic formulae embeddings support (Section~\ref{section:representation-subsystem})
\item examples of experiments with different RL algorithms (Section~\ref{section:experiments})
\item following the updated Gymnasium~\cite{towers_gymnasium_2023} API instead of the outdated OpenAI Gym
\end{itemize}

\texttt{gym-saturation} works with Python 3.8+. One can install it by \texttt{pip install gym-saturation} or \texttt{conda install -c conda-forge gym-saturation}. Then, provided Vampire and/or iProver binaries are on \texttt{PATH}, one can use it as any other Gymnasium environment:
\begin{minted}{python}
import gymnasium

import gym_saturation

# v0 here is a version of the environment class, not the prover
env = gymnasium.make("Vampire-v0")  # or "iProver-v0"
# edit and uncomment the following line to set a non-default problem
# env.set_task("a-TPTP-problem-path")
observation, info = env.reset()
print("Starting proof state:")
env.render()
# truncation means finishing an episode in a non-terminal state
# e.g. because of the externally imposed time limit
terminated, truncated = False, False
while not (terminated or truncated):
    # apply policy (e.g. a random available action)
    action = env.action_space.sample(mask=observation["action_mask"])
    print("Given clause:", observation["real_obs"][action])
    observation, reward, terminated, truncated, info = env.step(action)
print("Final proof state:")
env.render()
env.close()
\end{minted}                   
\section{Related work}\label{section:related-work}
Guiding provers with RL is a hot topic. Recent projects in this domain include TRAIL (Trial Reasoner for AI that Learns)~\cite{9669114}, FLoP (Finding Longer Proofs)~\cite{FLoP}, and lazyCoP~\cite{10.1007/978-3-030-86059-2_11}. We will now compare the new \texttt{gym-saturation} features with these three projects.

Usually, one guides either a new prover created for that purpose (lazyCoP; FLoP builds on fCoP~\cite{fCoP}, an OCaml rewrite of older leanCoP~\cite{OTTEN2003139}) or an experimental patched version of an existing one (TRAIL relies on a modified E~\cite{10.1007/978-3-030-29436-6_29}). Contrary to that, \texttt{gym-saturation} works with unmodified stable versions of Vampire~\cite{10.1007/978-3-642-39799-8_1} and iProver~\cite{DBLP:conf/cade/DuarteK20}.

In addition, known RL-guiding projects are prover-dependent: FLoP could, in principle, work with both fCoP and leanCoP but reported only fCoP experiments. TRAIL claims to be reasoner-agnostic, but to our best knowledge, no one has tried it with anything but a patched E version it uses by default. ~\cite{10.1007/978-3-030-86059-2_11} mentions an anonymous reviewer's suggestion to create a standalone tool for other existing systems, but we are not aware of further development in this direction. Quite the contrary, we have tested \texttt{gym-saturation} compatibility with two different provers (Vampire and iProver).

Deep learning models expect their input to be real-valued tensors and not, for example, character strings in the TPTP~\cite{DBLP:journals/jar/Sutcliffe17} language. Thus, one always uses a \emph{representation} (or \emph{embeddings}) --- a function mapping a (parsed) logic formula to a real vector. In lazyCoP and FLoP parts of embedding functions belong to the underlying provers, making it harder to vary and experiment with (e.g., one needs Rust or OCaml programming skills to do it). \texttt{gym-saturation} leaves the choice of representation open and supports any mapping from TPTP-formatted string to real vectors. The version described in this work also provides a couple of default options.
\section{Architecture and implementation details}\label{section:implementation}
\subsection{Architecture}
\texttt{gym-saturation} is compatible with Gymnasium~\cite{towers_gymnasium_2023}, a maintained fork of now-outdated OpenAI Gym standard of RL-environments, and passes all required environment checks. As a result of our migration to Gymnasium, its maintainers featured \texttt{gym-saturation} in a curated list of third-party environments~\footnote{\url{https://gymnasium.farama.org/environments/third_party_environments/}}.

Previously, \texttt{gym-saturation} guided an experimental pure Python prover~\cite{Shminke2022} which happened to be too slow and abandoned in favour of existing highly efficient provers: Vampire and iProver.

\begin{figure}
\includegraphics[width=\textwidth]{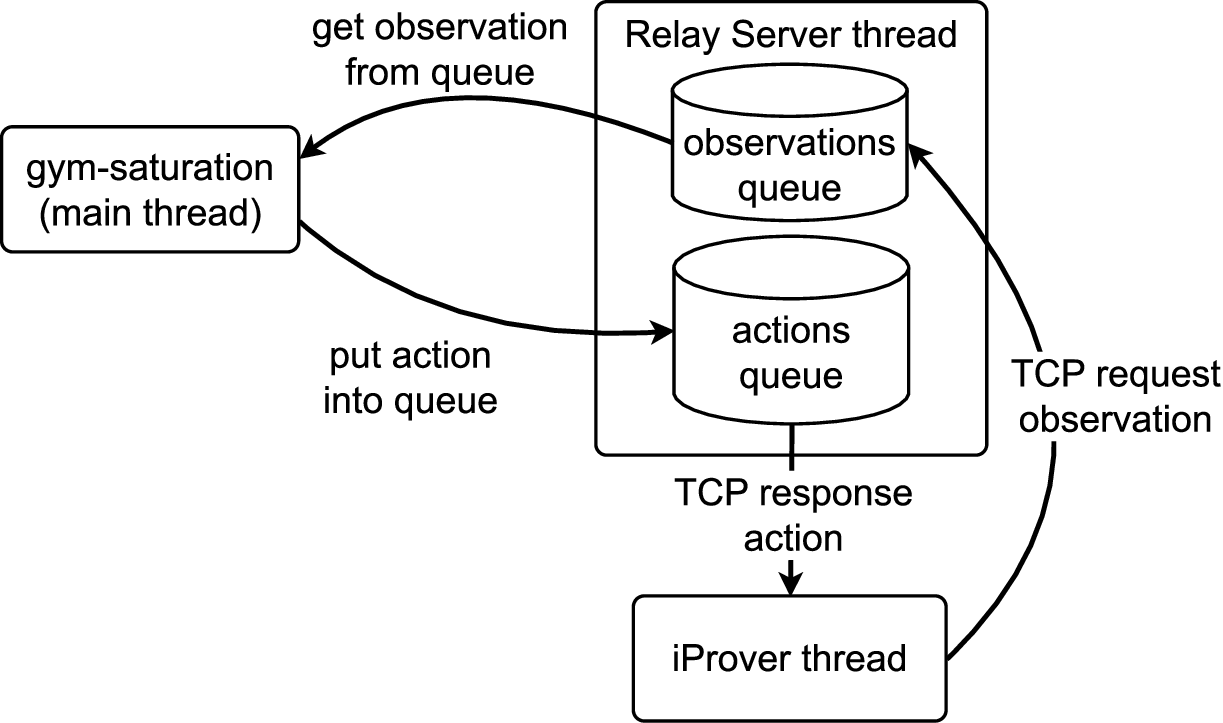}
\caption{\texttt{gym-saturation} interacting with iProver}\label{fig:iprover-gym}
\end{figure}

Although the \texttt{gym-saturation} user communicates with both iProver and Vampire in the same manner, under the hood, they use different protocols. For Vampire, we relied on the so-called manual (interactive) clause selection mode implemented several years ago for an unrelated task~\cite{10.1007/978-3-030-34968-4_28}. In this mode, Vampire interrupts the saturation loop and listens to standard input for a number of a given clause instead of applying heuristics. Independent of this mode, Vampire writes (or not, depending on the option \texttt{show\_all}) newly inferred clauses to its standard output. Using Python package \texttt{pexpect}, we attach to Vampire's standard input and output, pass the action chosen by the agent to the former and read observations from the latter. In manual clause selection mode, Vampire works like a server awaiting a request with an action to which it replies (exactly what an environment typically does).

iProver recently added support of being guided by external agents. An agent has to be a TCP server satisfying a particular API specification. So, iProver behaves as a client which sends a request with observations to some server and awaits a reply containing an action. To make it work with \texttt{gym-saturation}, we implemented a \emph{relay server}. It accepts a long-running TCP connection from a running iProver thread and stores its requests to a thread-safe queue, and pops a response to it from another such queue filled by \texttt{gym-saturation} thread. See Figure~\ref{fig:iprover-gym} for a communication scheme.
\subsection{Implementation details}
\subsubsection{Clause class}
A clause is a Python data class having the following keys and respective values:

\begin{itemize}
\item \texttt{literals} --- a string of clause literals in the TPTP format, e.g. \texttt{'member(X0,bb) | ~member(X0,b)'}
\item \texttt{label} --- a string label of a clause, e.g. `21'. Some provers (e.g. Vampire) use integer numbers for labelling clauses, but others (e.g. iProver) use an alphanumeric mixture (e.g. `c\_54')
\item \texttt{role} --- a string description of a clause role in a proof (hypothesis, negated conjecture, axiom, et cetera)
\item \texttt{inference\_rule} --- a string name of an inference rule used to produce the clause. It includes not only resolution and superposition but also values like `axiom' and `input' (for theorem assumptions)
\item \texttt{inference\_parents} --- a tuple of clause labels if needed by the inference rule (`axiom' doesn't need any, `factoring' expects only one, `resolution' --- two, et cetera)
\item \texttt{birth\_step} --- an integer step number when the clause appeared in the proof state. Axioms, assumptions, and the negated conjecture have birth step zero.
\end{itemize}

All these fields except the \texttt{birth\_step} (computed by the environment itself) are already available as separate entities (and not parts of TPTP-formatted strings) in iProver and Vampire output.
\pagebreak
\subsubsection{Environment class}
\paragraph{Observation} is a Python dictionary with several keys:
\begin{itemize}
\item \texttt{real\_obs} is a tuple of all clauses (processed and unprocessed). It can be transformed to tensor representation by so-called observation wrappers~\footnote{\url{https://gymnasium.farama.org/api/wrappers/observation_wrappers/}}. The \texttt{gym-saturation} provides several such wrappers for cases of external embeddings service or hand-coded feature extraction function
\item \texttt{action\_mask} is a numpy~\cite{harris2020array} array of the size \texttt{max\_clauses} (a parameter which one can set during the environment object instantiation) having a value $1.0$ at index $i$ if and only if a clause with a zero-based order number $i$ currently exists and can be a given clause (e.g. not eliminated as redundant). All other values of \texttt{action\_mask} are zeros. This array simplifies tensor operations on observation representations.
\end{itemize}
Limiting the total number of clauses in a proof state is a proxy of both random-access memory (each clause needs storage space) and time (a prover has to process each clause encountered) limits typical for the CASC~\cite{DBLP:journals/aicom/Sutcliffe21} competition. One can add a standard Gymnasium time-limit wrapper to limit the number of steps in an episode. Setting wall-clock time and RAM limits is not typical for RL research.
\paragraph{Action} is a zero-based order number of a clause from \texttt{real\_obs}. If a respective \texttt{action\_mask} is zero, an environment throws an exception during the execution of the \texttt{step} method. \paragraph{Reward} is $1.0$ after a step if we found the refutation at this step and $0.0$ otherwise. One can change this behaviour by either Gymnasium reward wrappers or by collecting trajectories in a local buffer and postprocessing them before feeding the trainer. \paragraph{Episode is terminated} when an empty clause \texttt{\$false} appears in the proof state or if there are no more available actions. \paragraph{Episode is truncated} when there are more than \texttt{max\_clauses} clauses in the proof state. Since the state is an (extendable) tuple, we don't raise an exception when a prover generates a few more clauses. \paragraph{Info} dictionary is always empty at every step by default. \paragraph{Render modes} of the environment include two standard ones (\texttt{'human'} and \texttt{'ansi'}), the first one printing and the second one returning the same TPTP formatted string.
\subsubsection{Multi-task environment}
The latest \texttt{gym-saturation} follows a Meta-World benchmark~\cite{pmlr-v100-yu20a} style and defines \texttt{set\_task} method with one argument --- a TPTP problem full path. If one resets an environment without explicitly setting a task in advance, the environment defaults to a simple group theory problem (any idempotent element equals the identity). Having a default task helps us keep compatibility with algorithms not aware of multi-task RL. One can inherit from \texttt{gym-saturation} environment classes to set a random problem at every reset or implement any other desirable behaviour.
\section{Representation subsystem}\label{section:representation-subsystem}
\subsection{Existing first-order formulae representations and related projects}
As mentioned in Section~\ref{section:related-work}, to apply any deep reinforcement learning algorithm, one needs a representation of the environment state in a tensor form first. There are many known feature engineering procedures. It can be as simple as clause age and weight~\cite{10.1007/978-3-030-29436-6_27}, or information extracted from a clause syntax tree~\cite{mockju-ecai20} or an inference lineage of a clause~\cite{10.1007/978-3-030-79876-5_31}. Representing logic formulae as such is an active research domain: for example, in~\cite{VectorRepresentations}, the authors proposed more than a dozen different embedding techniques based on formulae syntax. In communities other than automated deduction, researchers also study first-order formulae representation: for example, in~\cite{10.1007/978-3-031-21203-1_22}, the authors use semantics representation rather than syntax. One can also notice that first-order logic (FOL) is nothing more than a formal language, so abstract syntax trees of FOL are not, in principle, that different from those of programming language statements. And of course, encoding models for programming languages (like \texttt{code2vec}~\cite{alon2019code2vec} for Java) exist, as well as commercially available solutions as GPT-3~\cite{10.5555/3495724.3495883} generic code embeddings and comparable free models like LLaMA~\cite{DBLP:journals/corr/abs-2302-13971}.

To make the first step in this direction, we took advantage of existing pre-trained embedding models for programming languages and tried to apply them to a seemingly disconnected domain of automated provers.
\subsection{\texttt{ast2vec} and our contributions to it}
In~\cite{Paassen2022}, the authors proposed a particular neural network architecture they called \emph{Recursive Tree Grammar Autoencoders (RTG-AE)}, which encodes abstract syntax trees produced by a programming language parser into real vectors. Being interested in education applications, they also published the pre-trained model for Python~\cite{Paassen_McBroom_Jeffries_Koprinska_Yacef_2021}. To make use of it for our purpose, we furnished several technical improvements to their code (our contribution is freely available~\footnote{\url{https://gitlab.com/inpefess/ast2vec}}):
\begin{itemize}
\item a TorchServe~\cite{torchserve} handler for HTTP POST requests for embeddings
\item request caching with the Memcached server~\cite{memcached}
\item Docker container to start the whole subsystem easily on any operating system
\end{itemize}

\begin{figure}[H]
\includegraphics[width=\textwidth]{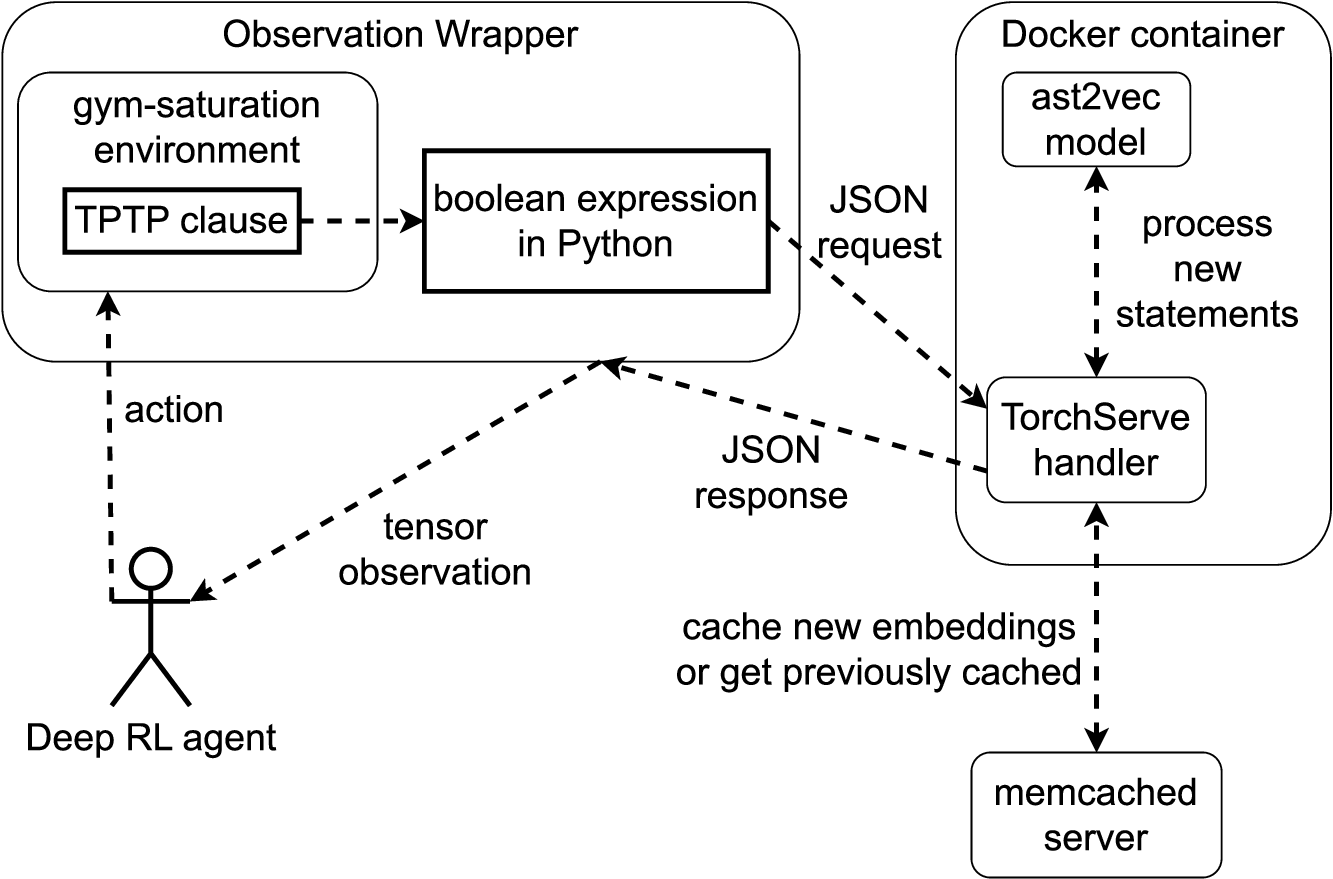}
\caption{gym-saturation communication with ast2vec} \label{fig:ast2vec}
\end{figure}

To integrate the \texttt{ast2vec} server with \texttt{gym-saturation} environments, we added Gymnasium observation wrappers, one of them mapping a clause in the TPTP language to a boolean-valued statement in Python (in particular, by replacing logic operation symbols, e.g. \texttt{=} in TPTP becomes \texttt{==} in Python). See Figure~\ref{fig:ast2vec} for a communication diagram. In principle, since a clause doesn't contain any quantifiers explicitly, one can rewrite it as a boolean-valued expression in many programming languages for which pre-trained embeddings might exist.

\subsection{Latency considerations}\label{subsection:latency-considerations}
Looking at Figure~\ref{fig:ast2vec}, one might wonder how efficient is such an architecture. The average response time observed in our experiments was $2ms$ (with a $150ms$ maximum). A typical natural language processing model which embeds whole texts has a latency from $40ms$ to more than $600ms$~\cite{nvidia-blog} (depending on the model complexity and the length of a text to embed) when run on CPU, so there is no reason to believe that \texttt{ast2vec} is too slow. When evaluating a prover, one usually fixes the time limit: for example, $60s$ is the default value for Vampire. Being written in C++ and with a cornucopia of optimisation tweaks, Vampire can generate around a million clauses during this relatively short timeframe. Thus, to be on par with Vampire, a representation service must have latency around $60\mu s$ (orders of magnitude faster than we have). There can be several ways to lower the latency:
\begin{itemize}
\item inference in batches (one should train the embedding model to do it; \texttt{ast2vec} doesn't do it out of the box). The improvement may vary
\item use GPU. NVIDIA reports around 20x improvement vs CPU~\cite{nlu-with-tensorrt-bert}. However, throwing more GPUs won't be as efficient without batch inference from the previous point
\item request an embedding for a binary object of an already parsed clause instead of a TPTP string. It means not repeating parsing already done by a prover, which might lower the latency substantially. To do this, one will have to patch an underlying prover to return binary objects instead of TPTP strings
\item use RPC (remote procedure call) instead of REST protocol. TorchServe relies on REST and parcels in JSON format, and in gRPC~\cite{grpc}, they prefer the binary \texttt{protobuf} format. One rarely expects sub-millisecond latency from REST, although for RPC, $150\mu s$ is not unusual. This point doesn't make much sense without the previous one
\end{itemize}

\section{Usage examples}\label{section:experiments}
We provide examples of experiments easily possible with \texttt{gym-saturation} as a supplementary code to this paper~\footnote{\url{https://github.com/inpefess/ray-prover/releases/tag/v0.0.3}}. We don't consider these experiments as being of any scientific significance per se, serving merely as illustrations and basic usage examples. Tweaking the RL algorithms' meta-parameters and deep neural network architectures is out of the scope of the present system description.

We coded these experiments in the Ray framework, which includes an RLlib --- a library of popular RL algorithms. The Ray is compatible with Tensorflow~\cite{tensorflow2015-whitepaper} and PyTorch~\cite{NEURIPS2019_bdbca288} deep learning frameworks, so it doesn't limit a potential \texttt{gym-saturation} user by one.

In the experiments, we try to solve \texttt{SET001-1} from the TPTP with \texttt{max\_clauses=20} (having no more than twenty clauses in the proof state) for guiding Vampire and \texttt{max\_clauses=15} for iProver. This difference is because even a random agent communicating to iProver manages to always solve \texttt{SET001-1} by generating no more than twenty clauses. We wanted training to start, but keep the examples as simple as possible, so we chose to harden the constraints instead of moving on to a more complicated problem.

In one experiment, we organise clauses in two priority queues (by age and weight) and use an action wrapper to map from a queue number ($0$ or $1$) to the clause number. That means we don't implant these queues inside provers but follow a Gymnasium idiomatic way to extend environments. Of course, Vampire and iProver have these particular queues as part of their implementation, but our illustration shows one could use any other priorities instead. It transforms our environment into a semblance of a 2-armed bandit, and we use Thompson sampling~\cite{pmlr-v28-agrawal13} to train. This experiment reflects ideas similar to those described in~\cite{10.1007/978-3-031-10769-6_38}.

In another experiment, we use \texttt{ast2vec} server for getting clause embeddings and train a Proximal Policy Optimisation (PPO) algorithm as implemented in the Ray RLlib. The default policy network there is a fully connected one, and we used $256\times20$ tensors as its input ($256$ is an embedding size in \texttt{ast2vec}, and $20$ is the maximal number of clauses we embed). So, the policy chooses a given clause given the embeddings of all clauses seen up to the current step (including those already chosen or judged to be redundant/subsumed). Such an approach is more similar to~\cite{FLoP}.

\begin{figure}[H]
\includegraphics[width=\textwidth]{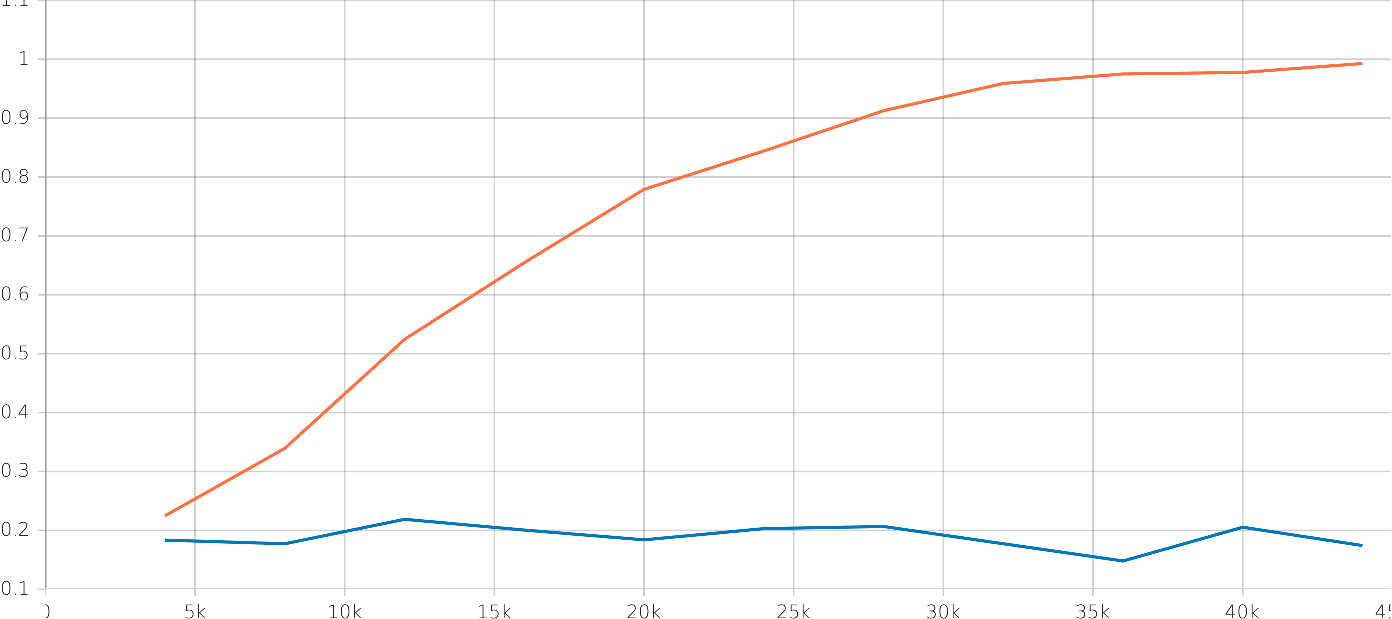}
\caption{Episode reward mean vs the total number of steps. The blue line is for a random agent and the orange one --- for the PPO. Both agents guide Vampire}\label{fig:mean-reward}
\end{figure}

We provide Figure~\ref{fig:mean-reward} as a typical training process chart.

\section{Conclusion and future work}
We contributed a new version of \texttt{gym-saturation}, which continued to be free and open-source software, easy to install and use while promising assistance in setting up experiments for RL research in the automated provers domain. In the new version, we enabled anyone interested to conduct experiments with RL algorithms independently of an underlying prover implementation. We also added the possibility of varying representations as external plug-ins for further experimentation. We hope that researchers having such an instrument can focus on more advanced questions, namely how to generate and prioritise training problems to better transfer search patterns learned on simpler theorems to harder ones.

Our experience with adding Vampire and iProver support to \texttt{gym-saturation} shows that working tightly with corresponding prover developers is not mandatory, although it might help immensely. Implementing the prover guidance through the standard I/O (as in Vampire) seems to be relatively easy, and we hope more provers will add similar functionality in future to be more ML-friendly. Such provers could then profit from using any other external guidance (see~\cite{LPAR2023:Guiding_an_Instantiation_Prover} for a different system using the same iProver technical features as we did).

We identify a discerning and computationally efficient representation service as a bottleneck for our approach and envision an upcoming project of creating a universal first-order logic embedding model usable not only by saturation-style provers but also tableaux-based ones, SMT-solvers, semantic reasoners, and beyond.
\subsubsection{Acknowledgements} We would like to thank Konstantin Korovin for the productive discussion and for adding the external agents' communication feature to iProver, without which this work won't be possible. We also thank anonymous reviewers for their meticulous suggestions on improving the present paper.

%
%
%
\bibliographystyle{splncs04}
\bibliography{gym-saturation}
\end{document}